\documentclass{preprint}
\usepackage{booktabs}
\usepackage{graphicx}
\usepackage[colorlinks=true]{hyperref}
\usepackage[numbers,comma,sort&compress]{natbib} 
\usepackage{xcolor}

\usepackage{enumitem}
\usepackage{siunitx}
\usepackage{makecell}

\title{CPGPrompt: Translating Clinical Guidelines into LLM-Executable Decision Support}
\author[1,$\dagger$]{Ruiqi Deng}
\author[2,3,$\dagger$]{Geoffrey Martin}
\author[4]{Tony Wang}
\author[5]{Gongbo Zhang}
\author[6]{Yi Liu}
\author[5]{Chunhua Weng}
\author[7,8,9]{Yanshan Wang}
\author[10,11]{Justin F Rousseau}
\author[3,12,*]{Yifan Peng}
\affil[1]{Information Science (Health Tech), Cornell Tech, New York, NY, USA}
\affil[2]{Systems Engineering, Cornell University, Ithaca, NY, USA}
\affil[3]{Population Health Sciences, Weill Cornell Medicine, New York, NY, USA}
\affil[4]{Computer and Information Science, Cornell University, Ithaca, NY, USA}
\affil[5]{Department of Biomedical Informatics, Columbia University, New York, NY, USA}
\affil[6]{Department of Medicine, Weill Cornell Medicine, New York, NY, USA}
\affil[7]{Department of Health Information Management, University of Pittsburgh, Pittsburgh, PA, USA}
\affil[8]{Clinical and Translational Science Institute, University of Pittsburgh, Pittsburgh, PA, USA}
\affil[9]{Department of Neurology, UT Southwestern Medical Center, Dallas, TX, USA}
\affil[10]{Peter O'Donnell Jr. Brain Institute, UT Southwestern Medical Center, Dallas, TX, USA}
\affil[11]{Clinical Informatics Center, University of Texas Southwestern Medical Center, Dallas, USA}
\affil[12]{Institute of Artificial Intelligence for Digital Health, Weill Cornell Medicine, New York, NY USA}
\affil[*]{Corresponding author(s). Email(s): \url{yip4002@med.cornell.edu}}
\affil[$\dagger$]{These authors contributed equally to this work.}

\begin{document}

\maketitle

\begin{abstract}
\textbf{Objective}: Clinical practice guidelines (CPGs) provide evidence-based recommendations for patient care; however, integrating them into Artificial Intelligence (AI) remains challenging. Previous approaches, such as rule-based systems or black-box AI models, face significant limitations, including poor interpretability, inconsistent adherence to guidelines, and narrow domain applicability. To address this, we develop and validate CPGPrompt, an auto-prompting system that converts narrative clinical guidelines into large language models (LLMs).

\textbf{Materials and Methods}: Our framework translates CPGs into structured decision trees and utilizes an LLM to dynamically navigate them for patient case evaluation. Synthetic vignettes were generated across three domains (headache, lower back pain, and prostate cancer) and distributed into four categories to test different decision scenarios. System performance was assessed on both binary specialty-referral decisions and fine-grained pathway-classification tasks.

\textbf{Results}: The binary specialty referral classification achieved consistently strong performance across all domains (F1: 0.85-1.00), with high recall (1.00 $\pm$ 0.00). In contrast, multi-class pathway assignment showed reduced performance, with domain-specific variations: headache (F1: 0.47), lower back pain (F1: 0.72), and prostate cancer (F1: 0.77).

\textbf{Discussion:} Domain-specific performance differences reflected the structure of each guideline. The headache guideline highlighted challenges with negation handling. The lower back pain guideline required temporal reasoning. In contrast, prostate cancer pathways benefited from quantifiable laboratory tests, resulting in more reliable decision-making.

\textbf{Conclusion}: CPGPrompt demonstrates generalizability across diverse clinical domains while maintaining high sensitivity for referral decisions. Its transparent, auditable framework enables the systematic identification of failure modes and provides advantages over black-box AI approaches. However, persistent challenges with subjective clinical assessments indicate a need for targeted improvements and greater clinical robustness.
\end{abstract}


\section{Introduction}

Clinical practice guidelines (CPGs) serve as the cornerstone of evidence-based medicine, providing systematically developed recommendations to standardize healthcare delivery and improve patient outcomes \cite{donaldson1991medicare}. These guidelines synthesize the best available evidence to guide clinical decision-making across diverse healthcare settings. However, translating complex, narrative guidelines into executable clinical decision support systems remains a significant challenge in healthcare informatics \cite{peleg2013computer-interpretable}.

Manual interpretation of clinical guidelines can be time-consuming, inconsistent, and particularly challenging for non-specialists in primary care settings, where diverse clinical presentations must be rapidly triaged and managed \cite{wang2023barriers}. Traditional rule-based clinical decision support systems require extensive manual programming and struggle to adapt to the nuanced language and complex decision structures found in modern clinical guidelines \cite{shortliffe2006biomedical}.

Prior studies have sought to bridge this gap by automatically converting guideline texts or flowcharts into decision-tree structures. For instance, MedDM constructs LLM-executable clinical guidance trees from guideline flowcharts \cite{li2023meddm}, and Text2MDT explores automated extraction of medical decision trees from narrative medical texts \cite{zhu2024text2mdt}. These works demonstrate the potential of structured representations for clinical reasoning; however, both still require intensive manual curation and verification to ensure logical consistency and completeness, and neither addresses transparent, auditable execution or generalizability across domains.

The emergence of AI agents, exemplified by large language models (LLMs), offers new opportunities to automate clinical reasoning tasks \cite{maity2025large, huang2024artificial}. Recent studies, such as Chen et al., have explored multi-agent systems designed to support clinical decision-making through coordinated reasoning and ethical oversight \cite{chen2025enhancing}. These models demonstrate strong capability in understanding natural language and generating contextually appropriate responses, suggesting potential applications in clinical decision support \cite{thirunavukarasu2023large}. However, direct application of LLMs to clinical decision-making raises concerns about reliability, interpretability, and the potential for hallucinations or inconsistent reasoning \cite{liu2023comprehensive}.

To address these challenges, we developed CPGPrompt, an auto-prompting AI agent that systematically converts narrative clinical guidelines or care-pathway diagrams into structured, LLM-executable guidance trees. These trees are then used to power chatbot-based execution engines that deliver guideline-adherent clinical decision support. Specifically, CPGPrompt consists of three components: practice guideline analysis, guidance tree construction, and an execution engine that traverses guidance trees through dynamic LLM prompting, resulting in a chatbot that presents the final decisions. Our approach addresses several limitations of existing clinical decision support systems, including translating complex narrative guidelines into executable logic, ensuring transparent and auditable decision-making processes, handling the nuanced language of clinical documentation (e.g., ambiguity or uncertainty expressed through modal or conditional phrasing), and generalizing across diverse clinical domains.

To evaluate the generalizability of CPGPrompt, we conducted a comprehensive assessment across three distinct clinical domains: headache management \cite{becker2015guideline}, lower back pain assessment \cite{top2015evidence}, and prostate cancer referral pathways \cite{young2015guideline}. Although the immediate focus is on these three domains, the proposed approach is disease-agnostic, and its broader adoption may significantly boost the secondary use of EHR data for translational clinical decision support analysis and quality metrics.

For each domain, we constructed synthetic patient vignettes simulating primary care notes with varied combinations of positive and negated clinical features. We used synthetic vignettes because they allowed us to systematically control guideline-relevant clinical features across domains. Our evaluation consisted of two tasks: a binary classification of referral versus non-referral cases, and a fine-grained classification that assigned the correct referral pathway in accordance with the guidelines. We demonstrated that CPGPrompt could successfully convert CPGs into structured, auditable decision trees and apply them across multiple clinical contexts. The framework achieved high sensitivity for referral decisions while maintaining interpretability and transparency, highlighting the potential of CPGPrompt as a generalizable and trustworthy approach to guideline-based clinical decision support.

\section{Materials and Methods}

\begin{figure}
\centering
\includegraphics[width=0.5\linewidth]{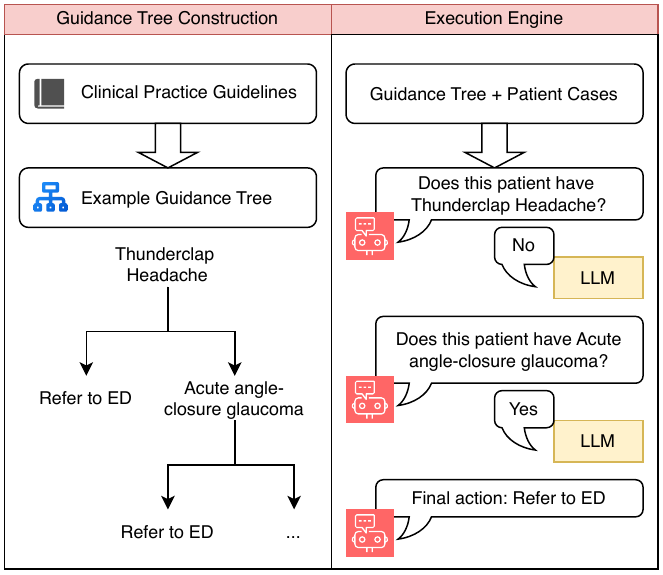}
\caption{Overview of CPGPrompt workflow. Clinical practice guidelines are first parsed into structured guidance trees, which are then combined with patient cases and traversed using LLM-based queries to produce referral or management actions.}
\label{fig:overview}
\end{figure}

CPGPrompt is designed with two agents (LLMs) that transform CPGs into a decision support artificial intelligence (AI) system (Figure \ref{fig:overview}). The first agent analyzes the guideline -- whether in narrative or diagrammatic form -- and constructs a structured guidance tree representing its diagnostic and referral logic. The second agent serves as a chatbot, using our designed execution engine to dynamically traverse this tree with context-aware prompts to evaluate patient vignettes. At each step, the system generates a transparent, auditable record of its reasoning while producing guideline-adherent management or referral decisions.

\subsection{Guideline Selection and Analysis}

We selected three CPGs representing diverse decision-making contexts. The headache guideline \cite{becker2015guideline} provides a structured approach for diagnosing and managing headaches in primary care, emphasizing the rapid identification of serious conditions that require urgent or emergent referral. The lower back pain guideline \cite{top2015evidence} introduces a multi-stage assessment process that incorporates red flags such as cauda equina syndrome symptoms, severe unremitting pain worse at night, or history of cancer, alongside timeline-based management features such as acute pain within 12 weeks versus persistent pain not improving after 1--6 weeks, distinguishing acute from chronic presentations. The prostate cancer guideline \cite{young2015guideline} integrates digital rectal examination findings, prostate-specific antigen lab test results, and age-based risk assessment to guide referral pathways. Together, these guidelines were chosen because they contain conditions that follow clearly defined urgent or emergent referral and management rules and capture different forms of clinical reasoning--immediate triage, longitudinal management, and test-based evaluation--providing a robust basis for assessing the generalizability of our framework.

We aimed to test CPGPrompt's ability to handle different CPG formats, including both narrative text and diagrammatic flowcharts. For the management of headaches and prostate cancer, we utilized text-based guidelines. We manually identified the sections describing diagnostic criteria, risk stratification, and referral recommendations. Relevant passages were extracted directly from the guideline documents and prepared as inputs for CPGPrompt. This process ensured that our framework faithfully reflected the original clinical logic while enabling downstream automation. For the lower back pain guideline, we used its official care-pathway diagram as the primary input, as it explicitly represents the clinical workflow through visual decision nodes, covering red-flag identification, timeline-based staging of symptom duration, and corresponding escalation steps for referral and management.

\subsection{Guidance Tree Construction}

\begin{figure}
\centering
\includegraphics[width=0.8\linewidth]{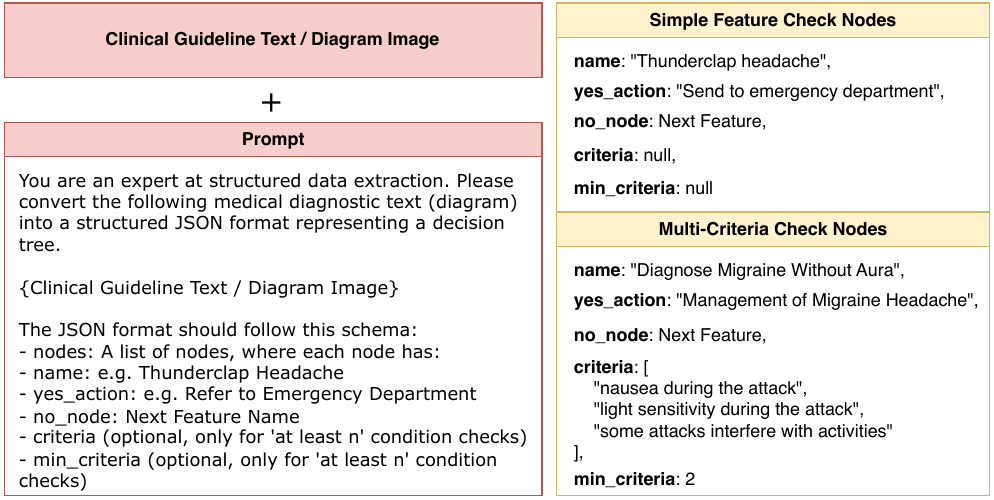}
\caption{Example prompt and output illustrating how CPGPrompt converts CPG into a structured JSON-based guidance tree, demonstrating both simple feature check nodes and multi-criteria check nodes.}
\label{fig:example}
\end{figure}

We used an LLM to automatically generate JSON-formatted guidance trees that capture the guideline's diagnostic and decision-making logic. The narrative guideline texts and diagram images were provided as inputs, using a structured prompting template (Figure \ref{fig:example}). The prompt instructed the model to extract decision logic into a standardized JSON schema, specifying node names, actions, and criteria. To ensure reliable performance, lengthy textual guidelines were first segmented into smaller, semantically coherent segments, and each segment was processed separately to generate a local subtree. These subtrees were subsequently merged into a unified decision tree representing the full guideline.

We identified two distinct types of decision nodes produced by the automated translation process:

\begin{enumerate}[nosep]
\item
  \textbf{Simple Feature Check Nodes} represent single-condition assessments, such as the presence of a ``thunderclap headache,'' that, when positive, immediately trigger a referral or management action.
\item
  \textbf{Multi-Criteria Check Nodes} represent more complex logic, requiring the simultaneous evaluation of multiple conditions. For example, the diagnostic criteria for migraine require that a patient meet at least 2 of 3 listed features before the diagnosis is confirmed.
\end{enumerate}

An example prompt and its corresponding JSON output for both node types are shown in Figure \ref{fig:example}.

Each automatically generated guidance tree was manually validated to ensure accurate representation of the source guideline's logic and decision pathways. Minor structural inconsistencies or logical errors identified during this process were addressed through prompt refinement and regeneration. Multi-criteria check nodes occasionally introduced small inaccuracies -- for example, the model sometimes miscounted the required number of criteria or loosely grouped related assessments within the same node. Temporal relationships and sequential dependencies were also less precisely extracted, as the model did not always treat time-based criteria (e.g., symptoms lasting more than x weeks) equivalently to clinical conditions. When generating from text, some feature names were imperfectly detected and required minor manual adjustment.

\subsection{Building a chatbot-based execution engine from the guidance tree}

We developed a chatbot-based execution engine that traverses the guidance tree to analyze patient cases. Starting from the root node, the engine evaluates each decision point, dynamically prompting the LLM with structured yes/no queries (e.g., ``Does the patient have a thunderclap headache?'').

For \textbf{Simple Feature Check Nodes}, the engine issues a single yes/no query for each clinical feature. A positive response triggers the predefined action (e.g., ``Refer to Emergency Department''), whereas a negative response redirects traversal to the specified next node.

For \textbf{Multi-Criteria Check Nodes}, the engine queries the LLM about each criterion independently. The responses are aggregated, and the node's threshold is applied to determine the outcome. If the minimum number of criteria is satisfied, the corresponding action is executed; otherwise, traversal continues along the ``no'' branch to the next node.

This process continues until a terminal node is reached, which either outputs a management or referral decision or signals the end of the decision pathway. Importantly, all queries, responses, and branching choices are recorded in a structured log. This ensures full transparency, enabling the reconstruction of the reasoning process for both clinical auditability and facilitating error analysis.

\subsection{Experimental Design}

\subsubsection{Synthetic Dataset Generation}

\begin{table}
\centering
\caption{Overview of the CPGPrompt benchmark. Dataset statistics across three clinical domains, including the number of vignettes, average vignette length, number of decision actions, referral distribution, and category breakdown.}
\label{tab:Overview}
\begin{tabular}{lrrr}
\toprule
& \textbf{Headache} & \textbf{Lower Back Pain} & \textbf{Prostate Cancer} \\
\midrule
\#Vignettes & 128 & 99 & 96 \\
Avg. Word Count & 140.5 & 239.3 & 136.2 \\
\#Actions & 24 & 7 & 11 \\
\multicolumn{4}{l}{%
\textbf{Vignette category}} \\
\;\;\;\;Single-criteria & 36 & 33 & 27 \\
\;\;\;\;Contrastive-criteria & 32 & 33 & 64 \\
\;\;\;\;Multi-criteria & 30 & 33 & NA \\
\;\;\;\;Exclusion-criteria & 30 & NA & 5 \\
\textbf{Binary classification} \\
\;\;Referral & 98 & 99 & 91 \\
\;\;Non-Referral & 30 & NA & 5 \\
\bottomrule
\end{tabular}
\end{table}

\begin{figure}
\centering
\includegraphics[width=\linewidth]{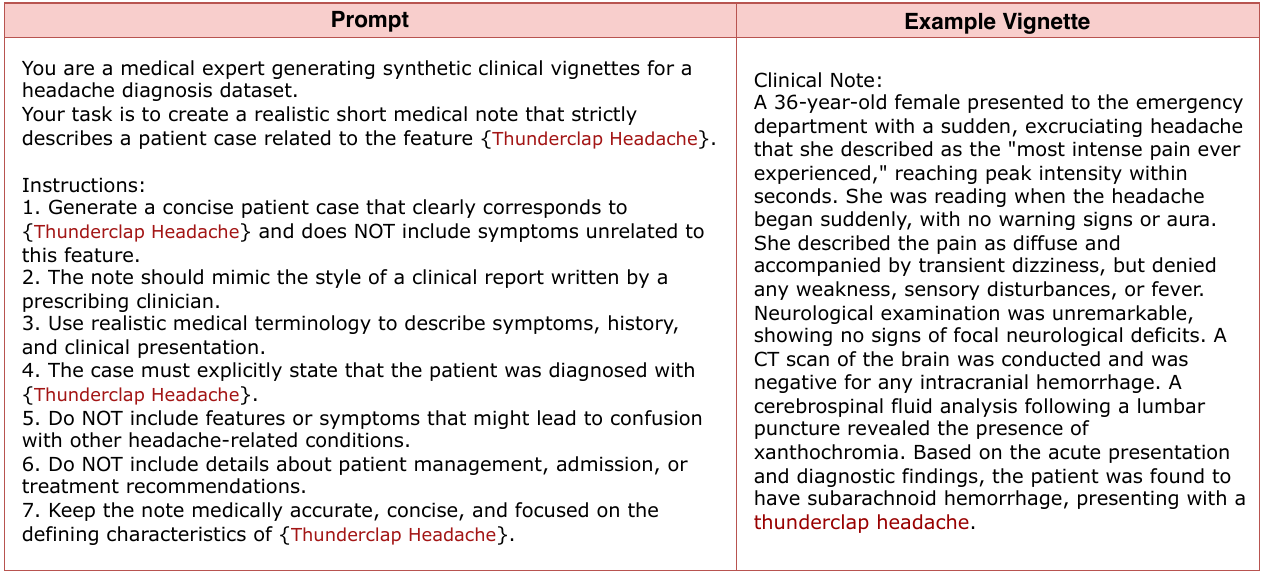}
\caption{Example of generating a single-criteria vignette in the headache domain. The figure shows the prompt template provided to the LLM (left) and a sample vignette output describing a patient with thunderclap headache (right).}
\label{fig:example single}
\end{figure}

To evaluate CPGPrompt, we used another LLM to generate synthetic clinical vignettes from the guidance tree, thereby creating controlled test scenarios across all three domains (Table \ref{tab:Overview}). Within each domain, vignettes were categorized into four categories based on the logical structure of the criterion in the guidelines. Each requires a set of conditions to be met. Please refer to Figure \ref{fig:example single} for an illustration of the prompt and an example vignette used in this process. Additional examples for all vignette categories are provided in Appendix \ref{sec:appendix}.

\begin{enumerate}[nosep]
\item
  \textbf{Single-criteria vignettes}: These vignettes include one clear positive condition. They are designed to test the system's basic diagnosis capabilities.
\item
  \textbf{Multi-criteria vignettes}: These vignettes feature multiple positive conditions, for example, ``Criteria A OR Criteria B''. They were designed to assess CPGPrompt's logic for prioritizing, sequencing, and managing competing diagnostic possibilities.
\item
  \textbf{Contrastive-criteria vignettes}: These vignettes explicitly exclude one clinical condition while including another, for example, ``(Criteria A and Criteria B) AND (NOT Criteria C)''. They are designed to test the system's ability to distinguish between similar conditions.
\item
  \textbf{Exclusion-criteria vignettes}: These vignettes do not include conditions of interest or fail to meet diagnostic criteria thresholds in the guidelines. They are intended to test the system's ability to avoid false positives and inappropriate referrals.
\end{enumerate}

Each vignette was generated to resemble realistic primary care clinical notes, including patient demographics, symptom descriptions, physical examination findings, and relevant medical history. To prevent bias in the evaluation, we also instructed the LLM to avoid providing explicit management recommendations in the vignettes. For the lower back pain guideline, we additionally created three vignette length conditions--short ($\leq 100$ words), medium (100-200 words), and long ($\geq 200$ words)--to examine the effect of vignette complexity on performance; this analysis is discussed later in the results.

We randomly selected 10 vignettes from each domain and had them reviewed by 3 clinicians for quality assurance. The clinicians confirmed that the vignettes were clinically coherent and guideline-consistent, requiring only minor wording adjustments; no major logical errors were identified.

\subsubsection{Evaluation framework}

For each clinical guideline, we conducted 5 independent test runs using GPT-4o with a temperature of 0.2 to balance consistency with natural variation in responses. Each run processed the full set of synthetic vignettes through the execution engine described in Section 2.3, which traverses the guidance trees by dynamically prompting the LLM and recording all intermediate steps. We evaluated CPGPrompt performance on two tasks:

\paragraph{Binary Classification.} Vignettes were categorized as either \emph{requiring referral} or \emph{not}. This task evaluates whether the system can reliably triage cases that need specialist intervention or escalation, regardless of the specific pathway. Therefore, true positives (TP) were cases in which the system correctly predicted a referral, false negatives (FN) were missed referrals, false positives (FP) were unnecessary referrals, and true negatives (TN) were correct non-referral predictions.

\paragraph{Multi-class Classification.} Each vignette was mapped to a single, guideline-defined referral pathway or management action based on its originating CPG. For example, in the headache domain, red-flag features necessitate emergency referral, whereas chronic primary headaches, such as migraines, are typically managed on a routine basis. The system outputs a single final action per vignette, determined by the chatbot traversing the guidance tree. For cases in the ``multi-criteria'' category, the highest-priority action among positive features was selected. In this task, TP, FP, and FN were computed per referral pathway or management action, and precision, recall, and F1 scores were averaged using macro-averaging to ensure equal weight of each class, regardless of class distribution.

\section{Results}

\subsection{Benchmark}

Table 1 summarizes our evaluation benchmark. It contains 128 vignettes on headache management, 99 on lower back pain, and 96 on prostate cancer. On average, vignette length differed across domains, with lower back pain being the longest (239 words), followed by headache (140 words) and prostate cancer (136 words). The number of unique clinical actions also varied, with 24 for headache, 7 for lower back pain, and 11 for prostate cancer. Vignettes were distributed across single-criteria, contrastive-criteria, multi-criteria, and exclusion-criteria categories. Referral and non-referral cases served as the binary labels for the binary classification task, while the set of clinical actions defined the classes for the multi-class task.

According to the headache management guidelines, we generated cases spanning all four categories. In contrast, the prostate cancer referral guidelines did not yield multi-criteria cases because most criteria are based on mutually exclusive laboratory tests (e.g., PSA $\geq 20$ \unit{\mu g/L} vs. $<20$ \unit{\mu g/L}), which cannot logically coexist within the same vignette. The lower back pain assessment contains no ``Exclusion-criteria'' cases, since all patients presenting with lower back pain necessarily exhibit at least one symptom that triggers a management step, ranging from simple analgesic treatment to ongoing monitoring, or referral for further evaluation. As a result, every lower back pain vignette was mapped to at least one clinical action, whereas headache or prostate cancer vignettes could, in some cases, be mapped to none.

This benchmark captures both the common and domain-specific challenges inherent to guideline-based reasoning. Beyond the category distribution, it also varies in terms of vignette length, the number of possible actions, and the ratio of referral to non-referral cases, offering a realistic and reproducible testbed for studying and comparing clinical decision support approaches.

\subsection{Overall Performance Across Domains}

Table \ref{tab:performance} presents model performance across the three guidelines. In the binary classification task, results were consistently strong, with F1 scores ranging from 0.90 to 1.00. Notably, recall was perfect in all domains, indicating that no cases requiring referral were missed. This represents a critical safety feature for LLM-based clinical decision support, as underdiagnosis, defined as falsely claiming that the patient is healthy, can lead to no clinical treatment when a patient needs it most. As a result, the patient will not receive much-needed attention in a timely manner. Underdiagnosis is potentially worse than misdiagnosis and overdiagnosis, because in the latter case, the patient still receives clinical and specialty evaluation and care, and the clinician can use other symptoms and data sources to clarify the mistake.

Performance in the multi-class classification task was lower, with F1 scores ranging from 0.44 to 0.77. While precision remained generally high, recall decreased, particularly in the cases involving headaches. This suggests that while the system reliably flagged cases that needed referral in the first place, it sometimes struggled to assign the correct referral pathway among multiple alternatives. To better understand these patterns, we next analyzed performance by vignette category.

\begin{table}
\centering
\caption{Overall Referral Classification Performance Across Clinical Domains}
\label{tab:performance}
\begin{tabular}{lccc}
\toprule
\textbf{Domain} & \textbf{Precision} & \textbf{Recall} & \textbf{F1-Score} \\
\midrule
\textbf{Binary Classification} \\
\;\;\;\;Headache & 0.82 $\pm$ 0.00 & 1.00 $\pm$ 0.00 & 0.90 $\pm$ 0.00 \\
\;\;\;\;Lower Back Pain & 1.00 $\pm$ 0.00 & 1.00 $\pm$ 0.00 & 1.00 $\pm$ 0.00 \\
\;\;\;\;Prostate Cancer & 1.00 $\pm$ 0.00 & 1.00 $\pm$ 0.00 & 1.00 $\pm$ 0.00 \\
\textbf{Multi-class Classification} \\
\;\;\;\;Headache & 0.66 $\pm$ 0.05 & 0.53 $\pm$ 0.02 & 0.44 $\pm$ 0.01 \\
\;\;\;\;Lower Back Pain & 0.80 $\pm$ 0.04 & 0.70 $\pm$ 0.03 & 0.72 $\pm$ 0.03 \\
\;\;\;\;Prostate Cancer & 0.84 $\pm$ 0.06 & 0.80 $\pm$ 0.04 & 0.77 $\pm$ 0.04 \\
\bottomrule
\end{tabular}
\end{table}

\subsection{Performance by Vignette Category}

Table \ref{tab:performance multi} presents multi-class results stratified by vignette category. The highest performance was observed in the ``Single-criteria'' and ``Multi-criteria'' cases, with particularly strong results for lower back pain (F1: 0.73 $\pm$ 0.03). These findings show that the system is well-suited to handle cases with clear positive clinical findings and can combine multiple criteria when underlying decision paths are explicitly defined.

Performance in the ``Exclusion-criteria'' category varied considerably across domains, highlighting how differences in guideline structure shape chatbot behavior. In the prostate cancer domain, ``Exclusion-criteria'' cases achieved the highest performance (F1: 1.0). This likely reflects the fact that prostate cancer criteria are primarily based on laboratory tests (e.g., PSA $\geq$ 20 \unit{\mu g/L} vs. $<20$ \unit{\mu g/L}), where negation is unambiguous, thereby facilitating accurate identification of true negatives. However, as this category comprised only five vignettes, the observed score should be interpreted with caution and viewed as indicative rather than definitive. By contrast, the headache guidelines use subjective symptom descriptions that are inherently less precise. In this context, the chatbot was observed to adopt a conservative decision-making strategy, frequently referring patients to avoid missing urgent conditions. While this behavior prioritizes safety, it also led to substantial over-referral, resulting in poor performance (F1: 0.06).

\begin{table}
\centering
\caption{Performance by Vignette Category (Multi-class Macro F1 Scores)}
\label{tab:performance multi}
\begin{tabular}{lrrr}
\toprule
\textbf{Category} & \textbf{Headache} & \textbf{Lower Back Pain} & \textbf{Prostate Cancer} \\
\midrule
Single-criteria & 0.51 $\pm$ 0.01 & 0.85 $\pm$ 0.02 & 0.76 $\pm$ 0.03 \\
Contrastive-criteria & 0.30 $\pm$ 0.02 & 0.79 $\pm$ 0.04 & 0.73 $\pm$ 0.03 \\
Multi-criteria & 0.41 $\pm$ 0.02 & 0.73 $\pm$ 0.03 & NA \\
Exclusion-criteria & 0.06 $\pm$ 0.01 & NA & 1.00 $\pm$ 0.00 \\
\bottomrule
\end{tabular}
\end{table}

\subsection{Traversal length analysis}

\begin{figure}
\centering
\includegraphics[width=0.8\linewidth]{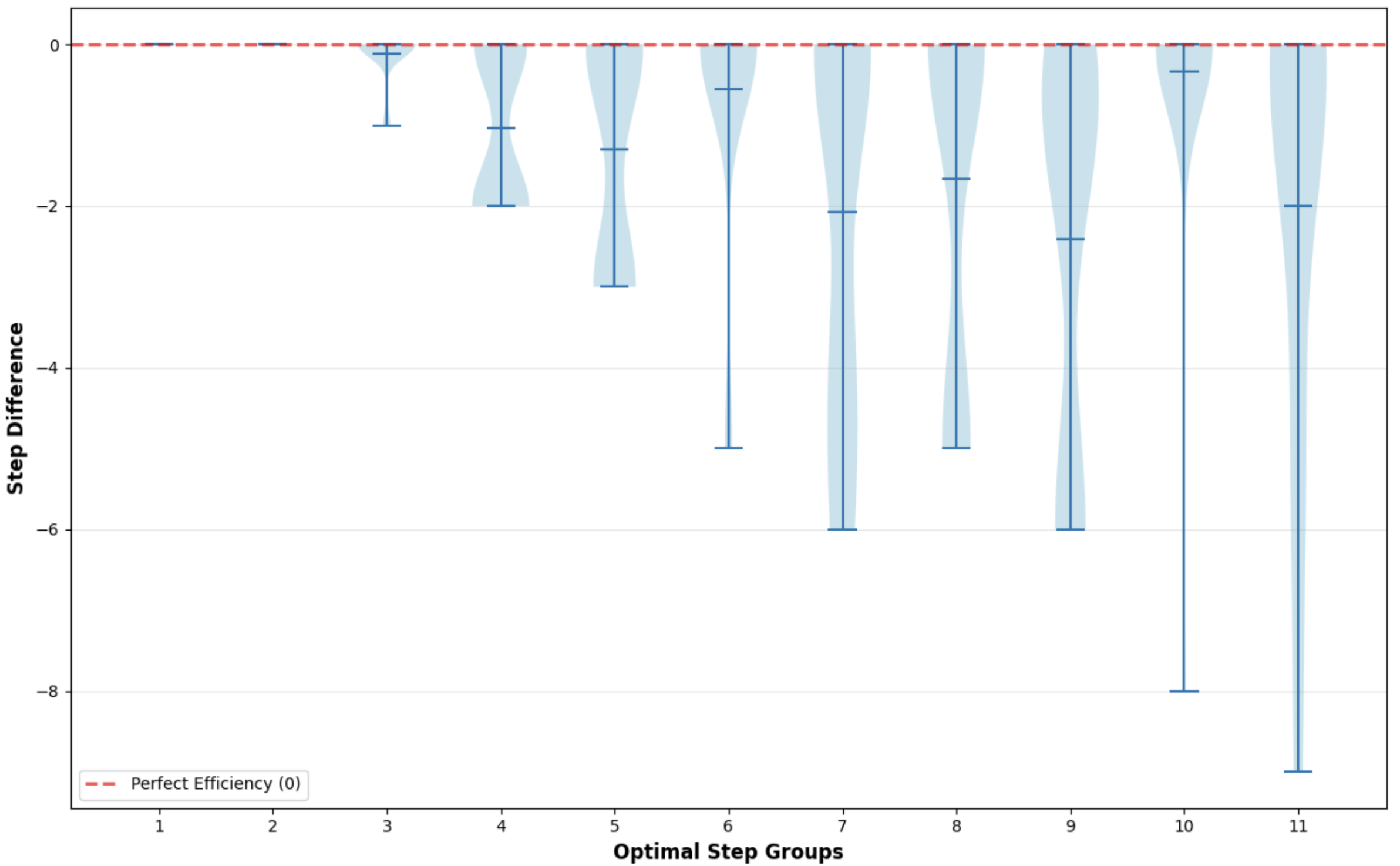}
\caption{\textbf{Distribution of step differences by optimal step group.} The x-axis groups cases by the number of steps required in the guideline-defined optimal path, while the y-axis shows the difference between predicted and optimal traversal lengths (predicted -- optimal). Negative values indicate early termination. Violin plots represent the distribution of differences, with horizontal bars showing the mean. The red dashed line at 0 marks perfect efficiency.}
\label{fig:distribution}
\end{figure}

In this experiment, we examined \textbf{traversal length} to understand how the chatbot progresses through the guidance trees. We define traversal difference as the difference between the number of steps taken by the system and the number of steps in the guideline-defined optimal path (predicted--optimal). Figure \ref{fig:distribution} shows that traversal differences are consistently negative, indicating the system tends to terminate earlier than the guideline-defined path.

This pattern suggests the model is strongly influenced by clinical criteria in early stages. Because decision trees are structured by urgency, earlier nodes represent more critical clinical criteria. As a result, the chatbot adopts a conservative strategy--preferring to trigger an action sooner rather than risk missing a relevant finding. Although this cautious behavior reduces traversal efficiency (fewer nodes are visited than along the guideline path), it substantially lowers the risk of misdiagnosis by ensuring that urgent cases are rarely missed. As discussed before, in clinical practice, this safety-oriented bias can be beneficial.

\subsection{Vignettes complexity level analysis}

To assess how vignette complexity affects CPGPrompt's performance, we evaluated the system on the lower back pain guideline across four vignette length conditions: unconstrained, long (minimum 200 words), short (maximum 100 words), and medium (between 100 and 200 words). These conditions varied the amount of clinical detail and narrative complexity while maintaining the same underlying diagnostic features. We present the results in Table \ref{tab:system} below.

\begin{table}
\centering
\caption{System performance by vignette complexity}
\label{tab:system}
\begin{tabular}{lrrrr}
\toprule
\textbf{Condition} & \textbf{Precision} & \textbf{Recall} & \textbf{F1-Score} & \makecell[tr]{\textbf{Mean Vignette}\\ \textbf{Length (Words)}} \\
\midrule
Unconstrained & 0.79 & 0.79 & 0.77 & 239 \\
Short & 0.88 & 0.86 & 0.84 & 76 \\
Medium & 0.79 & 0.80 & 0.76 & 153 \\
Long & 0.87 & 0.85 & 0.82 & 264 \\
\bottomrule
\end{tabular}
\end{table}

Short vignettes achieved the highest overall performance (F1: 0.84), followed closely by long vignettes (F1: 0.82). In contrast, unconstrained and medium-length vignettes showed reduced performance (F1: 0.77 and 0.76, respectively). This pattern suggests that well-structured vignettes with concise clinical details enable more reliable decision-making, whereas longer vignettes introduce noise into our system's decision-making process.

\subsection{Error analysis}

To better understand performance differences across domains, we examined systematic error patterns in relation to the specific characteristics of each guideline. Despite consistently high recall, domain-specific challenges persist and continue to shape system behavior.

For headache referrals, the most persistent challenge was negation handling. Many clinical findings mentioned in the vignettes described subjective symptoms, such as ``no red flag symptoms'' or ``denies neurological deficits''. CPGPrompt often defaulted to a cautious strategy that favored referral over the risk of a missed case. This tendency led to high recall but low performance in the ``Exclusion-criteria'' category.

For prostate cancer guidelines, negation exhibited fewer issues, since most decision points were based on quantifiable test thresholds (e.g., PSA $\geq 20$ \unit{\mu g/L} vs. $<20$ \unit{\mu g/L}). This accounts for the strong performance in the ``Exclusion-criteria'' cases, although integrating borderline test values with age-based risk factors introduced interpretive complexity.

For lower back pain, the primary difficulty lay in interpreting timeline-based clinical cues. The guideline requires distinguishing between acute versus persistent pain and identifying a lack of improvement within specific time windows. Typical decision points include phrases such as ``pain onset within 12 weeks -- first episode'' versus ``pain onset within 12 weeks -- not improving after 1--6 weeks.'' While human clinicians can infer these states from a single note, the model occasionally struggles to accurately parse and combine temporal phrases and map them to the relevant guideline branch.

Overall, these analyses demonstrate that CPGPrompt generalizes across diverse clinical practice guidelines; however, its limitations stem from the nature of the underlying features--subjective negations, quantified thresholds, or temporal assessments.

\section{Discussion}

In this study, we presented \textbf{CPGPrompt}, a generalizable auto-prompting AI agent framework that converts CPGs into structured, LLM-executable decision trees and integrates a chatbot to produce transparent, auditable clinical referral decisions. To systematically evaluate the framework, we constructed a new multi-domain benchmark of synthetic patient vignettes that captures diverse guideline structures, feature types, and reasoning challenges. Using this benchmark, we demonstrated the effectiveness of CPGPrompt across both binary referral classification and fine-grained multi-class pathway assignments, providing a detailed analysis of strengths, limitations, and areas for future improvement in LLM-based clinical decision support.

Our experiments demonstrate that CPGPrompt consistently shows strong recall across three clinical domains, a desirable safety characteristic for chatbots in patient triage. In healthcare settings, the risk of missing a referral-worthy case (false negative) is typically more serious than the risk of making an unnecessary referral (false positive) \cite{liu2023comprehensive}. Our findings suggest that CPGPrompt prioritizes this safety consideration while maintaining a reasonable level of precision.

Beyond domain-level performance, CPGPrompt provides a significant advantage over black-box AI approaches in clinical settings: transparency. Each decision step is logged and auditable, allowing clinicians to trace reasoning and identify potential errors. This transparency is crucial for clinical trust and regulatory approval \cite{london2019artificial}.

CPGPrompt also demonstrates generalizability. Despite domain-specific challenges, the framework consistently identified referral cases, suggesting scalability to additional CPGs with appropriate adaptation.

Compared to traditional rule-based systems, which require extensive manual programming and often fail to capture the nuanced language of guidelines \cite{shortliffe2006biomedical, papadopoulos2022systematic}, CPGPrompt leverages LLM capabilities with explicit, auditable decision pathways. Benefits include: (1) Clear, auditable reasoning steps. (2) Systematic handling of guideline logic and decision hierarchies. (3) Reduced risk of hallucination through structured prompting. However, our approach still requires manual verification of guidelines, which may limit its scalability compared to fully automated methods.

This study has several limitations. First, cases requiring clinical judgment, subjective decision-making, or integration of subtle clinical cues were consistently challenging. This limitation may be inherent to guideline-based approaches, suggesting the continued need for hybrid human-AI workflows rather than fully automated decision-making. Second, while the system's strong tendency toward referral supports patient safety, it also reflects a trade-off that led to over-referral in low-risk cases. In real-world settings, high sensitivity can come at the cost of unnecessary referrals, which may burden emergency services, increase healthcare costs, prolong wait times, and expose patients to avoidable testing or specialist consultations. Third, our study did not use real clinical notes--which often contain incomplete or noisy information--as our GPT model cannot be granted access to patient records due to privacy and security risks. Finally, assessment of patient progress, treatment response, and timeline-based decisions revealed challenges in processing temporal relationships and evolving clinical information. Future implementations should incorporate enhanced temporal reasoning capabilities, including the ability to integrate new symptoms or diagnostic data over time and adjust pathways accordingly.

Overall, our study shows that CPGPrompt can reliably apply narrative clinical guidelines in a structured and transparent manner. The framework strikes a balance between safety and interpretability, providing a foundation that can be extended to other domains. Future work should focus on strengthening areas such as temporal reasoning and uncertainty handling; however, the results presented here point to a practical path for bringing guideline-based decision support closer to clinical use.

\section*{Authorship contribution}

Ruiqi Deng and Geoffrey Martin: Writing -- original draft, Visualization, Investigation, Conceptualization. Tony Wang, Gongbo Zhang, Yi Liu, Chunhua Weng, Yanshan Wang, and Justin F Rousseau: Writing -- review \& editing. Yifan Peng: Writing -- review \& editing, Visualization, Supervision, Project administration, Funding acquisition, Conceptualization.

\section*{Declaration of competing interest}

The authors declare that they have no known competing financial interests or personal relationships that could have influenced the work reported in this paper.

\section*{Funding}

This research was supported by the National Library of Medicine under the grant numbers R01LM014306, R01LM014344, and R01LM014573.

\section*{Data Availability}

The data and code underlying this article are available on GitHub, at \url{https://github.com/bionlplab/CPGPrompt}.

\setlength{\bibsep}{3pt plus 0.3ex}
\bibliographystyle{unsrtnat}
\bibliography{ref}

\newpage
\appendix
\setcounter{subsection}{0}
\setcounter{table}{0}
\setcounter{figure}{0}
\renewcommand\figurename{Extended Data Figure} 
\renewcommand\tablename{Extended Data Table}
\renewcommand{\thesubsection}{S\arabic{subsection}}
\renewcommand{\thefigure}{S\arabic{figure}}
\renewcommand{\thetable}{S\arabic{table}}

\part*{Appendix}

\section{Representative Examples of All Vignette Categories}
\label{sec:appendix}

This appendix provides sample synthetic patient notes corresponding to each vignette category used in our evaluation framework: single-criteria, multi-criteria, contrastive-criteria, and exclusion-criteria. These examples illustrate how guideline-relevant features were systematically controlled in the synthetic dataset.

\subsection{Single-Criteria Vignette Example}

\begin{description}[nosep]
\item[Domain:] Headache 
\item[Positive Feature:] Thunderclap headache
\item[Vignette:] A 36-year-old female presented to the emergency department with a sudden, excruciating headache that she described as the "most intense pain ever experienced," reaching peak intensity within seconds. She was reading when the headache began suddenly, with no warning signs or aura. She described the pain as diffuse and accompanied by transient dizziness, but denied any weakness, sensory disturbances, or fever. Neurological examination was unremarkable, showing no signs of focal neurological deficits. A CT scan of the brain was conducted and was negative for any intracranial hemorrhage. A cerebrospinal fluid analysis following a lumbar puncture revealed the presence of xanthochromia. Based on the acute presentation and supporting diagnostic findings, the patient was diagnosed with a Thunderclap headache, with considerations of potential underlying etiology such as cerebral vasospasm or unruptured aneurysm.
\end{description}

\subsection{Multi-Criteria Vignette Example}

\begin{description}[nosep]
\item[Domain:] Lower Back Pain
\item[Positive Feature:] Widespread neurological signs, Pain onset within 12 weeks - first episode
\item[Vignette:] A 45-year-old male presents with acute onset lower back pain for the past 10 weeks, his first episode. He reports diffuse radiating pain to both lower extremities, accompanied by bilateral leg weakness and intermittent numbness extending to the feet. On examination, there is decreased strength (4/5) in dorsiflexion and plantarflexion bilaterally, diminished patellar and Achilles reflexes, and reduced sensation to light touch in a glove-and-stocking distribution. Straight leg raise test is positive bilaterally at 30 degrees. No prior history of back pain or neurological symptoms noted. Symptoms began following a strenuous lifting incident.
\end{description}

\subsection{Contrastive-Criteria Vignette Example}

\begin{description}[nosep]
\item[Domain:] Prostate Cancer 
\item[Positive Feature:] Incidental elevated PSA results, PSA level is greater than or equal to 20 micrograms per liter
\item[Negative Features:] Suspicious low back pain such as that associated with reproducible percussion tenderness
\item[Vignette:] A 65-year-old male presented for a routine annual check-up without specific complaints. The patient explicitly denied experiencing any low back pain, and physical examination revealed no spinal discomfort or tenderness upon percussion. Routine blood tests showed an incidental finding of elevated prostate-specific antigen (PSA) at 22 micrograms per liter, significantly higher than the age-adjusted reference range. He has no previous history of prostate disease, and digital rectal examination was unremarkable. Further diagnostic assessment of the prostate was recommended based on these laboratory findings.
\end{description}

\subsection{Exclusion-Criteria Vignette Example}

\begin{description}[nosep]
\item[Domain:] Headache 
\item[Negative Features:] Unexplained Focal Signs, Late Onset Headache
\item[Vignette:] A 55-year-old female presented for an annual routine examination with no specific complaints reported. During the review of systems, she explicitly denied experiencing any unexplained focal neurological signs, such as unilateral weakness, sensory disturbances, or coordination issues. Furthermore, the patient also expressly denied the occurrence of any newly onset headaches or changes in headache patterns typically associated with later life, such as increased frequency or intensity. On clinical examination, the patient's neurological assessment revealed intact cranial nerve function, equal and reactive pupils, symmetrical deep tendon reflexes, and a steady gait. No focal deficits were observed, and overall, the neurological status was unremarkable. Her vital signs were within normal limits, and there was no indication of any acute or chronic headache-related conditions. Considering the patient's clear denial of symptoms and normal findings, no further neurological investigations were initiated.
\end{description}

\end{document}